\documentclass[letterpaper, 10 pt, conference]{ieeeconf}  

\usepackage{amsmath}
\DeclareMathOperator*{\argmin}{arg\,min}

\usepackage{graphicx}


\makeatletter

\newcommand{\Rmnum}[1]{\expandafter\@slowromancap\romannumeral #1@}
\makeatother

\usepackage{comment}
\usepackage{subfig}
\usepackage{graphicx}
\usepackage{siunitx}

\usepackage{array}

\usepackage{algorithm}

\usepackage{algorithmic}

\usepackage{ragged2e}

\usepackage{url}

\usepackage[english]{babel}

\newtheorem{theorem}{Theorem}
\newtheorem{corollary}{Corollary}

\begin{document}
\title{\LARGE \bf
SPARSE KERNEL PCA FOR OUTLIER DETECTION
}

\author{Rudrajit Das$^{*}$, Aditya Golatkar$^{*}$ and Suyash Awate\\
Indian Institute of Technology Bombay, Mumbai, India\\
* \small denotes equal contribution \vspace{-0.6cm}
}

\maketitle
\thispagestyle{empty}
\pagestyle{empty}

\begin{abstract}

In this paper, we propose a new method to perform Sparse Kernel Principal Component Analysis (SKPCA) and also mathematically analyze the validity of SKPCA. We formulate SKPCA as a constrained optimization problem with elastic net regularization (Hastie \textit{et al}.) in kernel feature space and solve it. We consider outlier detection (where KPCA is employed) as an application for SKPCA, using the RBF kernel. 
We test it on 5 real world datasets and show that by using just 4\% (or even less) of the principal components (PCs), where each PC has on average less than 12\% non-zero elements in the worst case among all 5 datasets, we are able to nearly match and in 3 datasets even outperform KPCA. We also compare the performance of our method with a recently proposed method for SKPCA by Wang \textit{et al}., and show that our method performs better in terms of both accuracy and sparsity.
We also provide a novel probabilistic proof to justify the existence of sparse solutions for KPCA using the RBF kernel. To the best of our knowledge, this is the first attempt at theoretically analyzing the validity of SKPCA.
\end{abstract}


\section{INTRODUCTION}
Kernel PCA (KPCA) is a non-linear version of PCA proposed in [7].
A major limitation of KPCA is that the eigen vectors of the covariance matrix in the kernel space are linear combinations of all the training data points, which becomes cumbersome for storage as well as querying a new test point.
Obtaining sparse coefficients in KPCA is of paramount importance for real world applications. This problem was first addressed in [8], where sparse kernel feature analysis method was proposed to obtain sparse coefficients, by imposing a $l_1$ penalty on the coefficients. Tipping in [9] proposed a Probabilistic PCA([10]) based method for SKPCA. However, such a method is data dependent as it does not ensure sparse solutions for all types of data. 
Achlioptas \textit{et al.} in [11] propose a sparse unbiased and low variance estimator for the eigenvectors using concentration inequalities.
Recently, Wang \textit{et al}. in [3] formulated SKPCA as a regression problem and solved it using Alternating Direction Method of Multipliers (ADMM). 


Outlier detection is a one class classification problem with possible multiple sub classes within the outlier class. The trained model for this task should be able to distinguish the learned inlier class from an outlier or novel class. One class SVM based approach in [5] and SVDD approach in [6] have previously been used for outlier detection. Another approach for outlier detection has been presented in [2], in which KPCA has been utilized for modeling the training set. 



The main contributions of this paper are two fold.
Firstly, we extend the work done in [1] to the case of SKPCA. We apply their technique in the kernel space to obtain a generalized version of the problem in Theorem 4 of [1] and also present a method to solve it. To the best of our knowledge, this method for SKPCA has not been employed before.
To validate our method, we consider the problem of outlier detection for which KPCA has been used before in [2]. By performing extensive experimentation on 5 real world datasets, we show that it performs nearly as well as KPCA (even better than KPCA for 3 datasets) and better than the recently proposed sparse KPCA approach in [3] as well as naive thresholding (simply picking the largest magnitude coefficients). It must be emphasized here that we are not claiming that our SKPCA method performs better than normal KPCA, instead we are merely claiming that it works almost as well as normal KPCA even under high sparsity.
Secondly, we provide a novel probabilistic proof to justify the existence of sparse solutions to the KPCA problem using the RBF kernel. To the best of our knowledge, this is the first attempt at mathematically justifying the validity of sparsifying the KPCA problem.


The paper is organized as follows - Section \Rmnum{2} gives a brief overview of the outlier detection algorithm in [2] using KPCA. Section \Rmnum{3} describes our method of Sparse KPCA for outlier detection. Section \Rmnum{4} presents our theoretical justification of sparse KPCA for the RBF kernel. Section \Rmnum{5} discusses and shows the results of our experiments. Section \Rmnum{6} concludes the paper.

\section{KPCA for Outlier Detection}
Hoffmann in [2] presented an outlier detection approach using KPCA. In [2], spherical potential of a point $z$ (mapped to $\phi(z)$ in feature space) is defined with respect to a data distribution as 
$p_s(z) = ||\phi(z) - \phi_0||^{2}$, where $\phi_0$ is the mean of the data in feature space. The reconstruction error is defined as $p(\widetilde{\phi}(z)) = \langle\widetilde{\phi}(z),\widetilde{\phi}(z) \rangle - \langle W\widetilde{\phi}(z),W\widetilde{\phi}(z) \rangle$, where $\widetilde{\phi}(z) = \phi(z) - \phi_0$ and $W$ contains the top $q$ eigen vectors of the data in kernel space, arranged along the rows. We seek to get a sparse representation of $W$ using our SKPCA approach. Specifically, if the reconstruction error of a point 
is less than a certain threshold, then it is classified as an inlier otherwise as an outlier. 
When we use an RBF kernel, inliers lie inside a sphere whereas outliers lie outside it.

\setcounter{equation}{0}

\section{Sparse KPCA For Outlier Detection}
In this section, we explain our method of obtaining sparse coefficients for the PCs in kernel space and its application in the outlier detection problem. 
\subsection{Sparse KPCA (SKPCA) Algorithm}
We have $n$ data points $\{x_{1},x_{2},\ldots,x_{n}\}$ which get mapped to $\{{\phi}(x_{1}),{\phi}(x_{2}),\ldots,{\phi}(x_{n})\}$ in the kernel space. We assume that the data in the kernel space is centered. Similar to [1], we first propose the following regression problem to obtain the coefficients of the first PC (${\alpha_1}$):
\begin{equation}
\begin{aligned}
& \underset{\alpha,\beta}{\text{min }}
\mathrm \sum_{k=1}^n (\|{\phi}(x_{k}) - v_{1}\langle v_{2},{\phi}(x_{k}) \rangle\|^2
+{\lambda}{\beta_{1,k}}^2)
\end{aligned}
\end{equation}
\vspace{-0.4cm}
\begin{equation*} 
\begin{aligned}
\vspace{-0.4cm}
\text{where } v_{1} = \sum_{k=1}^n {\alpha_{1,k}}{\phi}(x_{k}) \text{ , } v_{2} = \sum_{k=1}^n {\beta_{1,k}}{\phi}(x_{k})
\end{aligned}
\end{equation*}
\begin{equation*} 
\begin{aligned}
\vspace{-0.55cm}
\text{such that } \langle v_{1},v_{1} \rangle = 1 \implies\ {\alpha_1}^{T}K{\alpha_1} = 1
\end{aligned}
\end{equation*}

Here, $K$ is the Gram matrix of the centered data~\cite{c7}. Also, $\alpha_1$ and $\beta_1$ are $n$ x $1$ vectors with $\alpha_{1,i}$ and $\beta_{1,i}$ as their $i^{th}$ elements respectively. So, we must minimize equation (1) (denote it by $J$) subject to ${\alpha_1}^{T}K{\alpha_1} = 1$. Substituting the values of $v_{1}$ and $v_{2}$ in $J$, we get:
\begin{equation}
\begin{aligned}
&  J = \ \sum_{k=1}^n \|{\phi}(x_{k}) - (\sum_{i=1}^n {\alpha_{1,i}}{\phi}(x_{i})) {\overline{\beta_{1,k}}}\|^2 +{\lambda}\sum_{k=1}^n {\beta_{1,k}}^2
\end{aligned}
\end{equation}
\vspace{-0.2in}
\begin{equation*}
\begin{aligned}
\text{where } {\overline{\beta_{1,k}}} = \sum_{i=1}^n {\beta_{1,i}}K(x_{k},x_{i}) = \text{$k^{th}$ element of } K{\beta_1}
\end{aligned}
\end{equation*}



\vspace{-0.2cm}
Rewriting (2) in matrix form and using ${\alpha_1}^{T}K{\alpha_1} = 1$, $K^{T}=K$, i.e. the Gram matrix is symmetric, we get:
\begin{equation}
\begin{aligned}
&  J = tr(K) - 2{\alpha_1}^{T}K^2{\beta_1} + {\beta_1}^{T}K^2{\beta_1} + \lambda{\beta_1}^{T}{\beta_1}
\end{aligned}
\end{equation}


Observe that this is the same as the PCA problem in [1] with the data matrix $X$ replace by the Gram matrix $K$. Even though the optimization problem is exactly the same, the constraint is different, namely ${\alpha_1}^{T}K{\alpha_1} = 1$ instead of ${\alpha_1}^{T}{\alpha_1} = 1$. This is quite intuitive.

Partially differentiating with respect to ${\beta}$ and setting it to $0$ :
\vspace{-0.03in}
\begin{equation}
\begin{aligned}
\frac{\partial J}{\partial \beta_1} = 0
\implies\ \widehat{\beta_1} = (K^2+\lambda I)^{-1} K^2{\alpha_1}
\end{aligned}
\end{equation}


Substituting $\widehat{\beta_1}$ from (4) in $J$, we get:
\vspace{-0.02in}
\begin{equation}
\begin{aligned}
J = tr(K) - {\alpha_1}^{T}K^2(K^2+\lambda I)^{-1}K^2{\alpha_1}\\
\end{aligned}
\end{equation}

Now, minimizing $J$ with respect to $\alpha_1$ subject to $\alpha_{1}^{T}K\alpha_1 = 1$ is equivalent to the following problem:
\begin{equation}
\begin{aligned}
\widehat{\alpha_1} = \underset{\alpha_1}{\operatorname{argmax}}{\hspace{0.1in}} 
{\alpha_1}^{T}K^2(K^2+\lambda 
I)^{-1}K^2{\alpha_1} 
\end{aligned}
\end{equation}

Let $H$ denote the Lagrangian cost function obtained by imposing the equality constraint and $\rho$ be the Lagrange multiplier.
\begin{equation}
\begin{aligned}
H = {\alpha_1}^{T}K^{2}(K^{2}+{\lambda}I)^{-1}K^{2}{\alpha_1}-\rho({\alpha_1}^{T}K{\alpha_1}-1)
\end{aligned}
\end{equation}
\vspace{-0.3cm}
\begin{equation}
\begin{aligned}
\frac{\partial H}{\partial \alpha_1} = 0
\implies\ K(K^{2}+{\lambda}I)^{-1}K^{2}{\alpha_1}=\rho {\alpha_1}
\end{aligned}
\end{equation}

Now, let $K=EDE^{T}$ be the eigen decomposition of $K$.
\begin{equation}
\implies\ K(K^{2}+{\lambda}I)^{-1}K^{2}
= E\frac{D^3}{(D^2+\lambda I)}E^{T}
\end{equation}

For the sparse PCA problem in [1], we had $D^4$ in the numerator above and not $D^3$. However, $\frac{D^3}{(D^2+\lambda I)}$ is also monotonically increasing and so the value of the objective function will be maximum when $\widehat{\alpha_1}$ is proportional to the eigen vector corresponding to the largest eigen value (as was the case for the sparse PCA problem in [1]), i.e. $\widehat{\alpha_1} = \pm (\frac{E_1}{\sqrt{D_1}})$ where $D_1$ is the maximum eigen value and $E_1$ is the corresponding eigen vector and $ \widehat{\beta_1} = (K^{2}+\lambda I)^{-1}K^2 \widehat{\alpha_1}=(\frac{D_1^2}{D_1^2+ \lambda})(\frac{E_1}{\sqrt{D_1}})$.

This was only for the first PC. Now, we will deal with the case of the first $m$ PCs and introduce the Lasso term as done in [1] so as to obtain sparse solutions. 
For this, we must minimize the following objective function with respect to $\alpha_m$ and $\beta_m$ which are $n$ x $m$ matrices:
\begin{multline}
\begin{aligned}
J = tr(K) - 2tr({\alpha_m}^{T}K^2{\beta_m}) + tr({\beta_m}^{T}(K^2+\\ \lambda I){\beta_m}) +\sum_{j=1}^m \lambda_j |\beta_{m}^{j}|_{1} \text{ subject to } \alpha_m^{T}K\alpha_m = I_{m} 
\end{aligned}
\end{multline}

Here, $\beta_{m}^{j}$ denotes the $j^{th}$ column in $\beta_m$ and $I_{m}$ denotes the $m$ x $m$ identity matrix.

Note that this is a convex problem with respect to $\alpha_m$ and $\beta_m$ individually and so we shall follow an 
alternating optimization scheme, 
i.e. keeping $\alpha_m$ fixed, find the optimal $\beta_m$, then keeping $\beta_m$ fixed, find the optimal $\alpha_m$ and repeat this process until convergence.

The solution for the first step of this iterative approach for sparse KPCA, i.e. keeping $\alpha_m$ fixed and finding the optimal $\beta_m$ is the same as that for sparse PCA in [1]. Specifically, $\beta_{m}^{j}$ is obtained by solving the following objective function:
\begin{equation}
\begin{aligned}
\beta_{m}^{j}=\text{}
& \underset{\beta^{*}}{\text{argmin}}
& \mathrm {\beta}^{*T}(K^2+\lambda I){\beta}^{*}- 2{\alpha_{m}^{j}}^{T}K^2{\beta}^{*}+\lambda_j|\beta^{*}|_{1}
\end{aligned}
\end{equation}
Note that $\alpha_{m}^{j}$ here refers to the $j^{th}$ column of $\alpha_{m}$ which was obtained in the previous iteration. This is the familiar naive elastic net problem [4] in ${\beta^{*}}$, which can be solved using the LARS-EN algorithm [4].

The solution for the second step of this iterative approach for sparse KPCA, i.e. keeping $\beta_{m}$ fixed and finding the optimal $\alpha_{m}$ is different as compared to that for sparse PCA in [1]. Specifically, we must solve the following problem:
\begin{equation}
\begin{aligned}
\alpha_{m} = \text{}
& \underset{\alpha^{*}}{\text{argmax}}
& \mathrm tr({\alpha^{*}}^{T}K^2{\beta_m})
\text{ subject to } {\alpha^{*}}^{T}K{\alpha^{*}} = I_{m} 
\end{aligned}
\end{equation}

Note that in sparse PCA, the objective function for finding the optimal $\alpha_m$ was the same but the constraint was different, namely $\alpha_m^{T}\alpha_m = I_{m}$ .

We present a corollary to Theorem 4 of [1] which enables us to solve the aforementioned problem in (12). We first restate Theorem 4 of [1] (renamed as \textit{Theorem 1} here) for ready reference of the reader followed by the corollary (\textit{Corollary 1}).
\begin{theorem}
Consider the following problem where $\alpha$, $\beta$ are $n$ x $m$ matrices ($n$ $>$ $m$):
\begin{equation*}
\begin{aligned}
\widehat{\alpha} = \text{ }
& \underset{\alpha}{\text{argmax}}
& \mathrm tr({\alpha}^{T}{\beta})
\text{ subject to } \alpha^{T}\alpha = I_{m} 
\end{aligned}
\end{equation*}
The solution to this problem is given by $\widehat{\alpha}=U{V}^{T}
\text{ where the SVD of } \beta = U{\Sigma}{V}^{T}$. 
\end{theorem}

\begin{corollary}
Consider the following problem where $\alpha$, $\beta$ are $n$ x $m$ matrices ($n$ $>$ $m$) and $Q$ is a $n$ x $n$ symmetric positive definite matrix:
\begin{equation*}
\begin{aligned}
\widehat{\alpha} = \text{ }
& \underset{\alpha}{\text{argmax}}
& \mathrm tr({\alpha}^{T}{\beta})
\text{ subject to } \alpha^{T}Q\alpha = I_{m} 
\end{aligned}
\end{equation*}
The solution to this problem is given by $\widehat{\alpha} = U\Sigma^{-1/2}U^{*}{V^{*}}^{T}$ where the SVD of $\Sigma^{-1/2}U^{T}{\beta} = U^{*}{\Sigma^{*}}{V^{*}}^{T}$ and the SVD of $Q = U\Sigma U^{T}$.
\end{corollary}



Note that \textit{Corollary 1} can be interpreted as an extension of \textit{Theorem 1} to a generalized inner product space (defined by the matrix $Q$).

We now present a proof for \textit{Corollary 1}. Let the SVD decomposition of $Q$ be $U\Sigma U^{T}$. Thus, $\alpha^{T}Q\alpha = (\alpha^{T}U\Sigma^{1/2})(\Sigma^{1/2}U^{T}\alpha)$, where $\Sigma^{1/2}$ is a diagonal matrix with diagonal entries equal to square roots of the corresponding entries of $\Sigma$. Now, denote $\alpha^{*} = (\Sigma^{1/2}U^{T}\alpha)$ and from the constraint, we must have ${\alpha^{*}}^{T}{\alpha^{*}} = I_{m}$. Also, $\alpha = (U\Sigma^{-1/2}{\alpha^{*}})$ and so ${\alpha}^{T}{\beta} = {{\alpha}^{*}}^{T}(\Sigma^{-1/2}U^{T}{\beta}) = {{\alpha}^{*}}^{T}{\beta^{*}}$, where ${\beta^{*}} = \Sigma^{-1/2}U^{T}{\beta}$. Thus, we have reduced the generalized problem to the following equivalent problem:
\begin{equation*}
\begin{aligned}
\widehat{{\alpha^{*}}} = \text{ }
& \underset{\alpha^{*}}{\text{argmax}}
& \mathrm tr({{\alpha^{*}}}^{T}{{\beta^{*}}})
\text{ subject to } {{\alpha^{*}}}^{T}{\alpha^{*}} = I_{m} 
\end{aligned}
\end{equation*}
This is exactly the problem whose solution has been given in Theorem 4 of [1] (\textit{Theorem 1} in this paper). We directly use that result to get:
\begin{equation*}
\begin{aligned}
\widehat{{\alpha^{*}}} = U^{*}{V^{*}}^{T}
\text{, SVD of } {{\beta^{*}}} = U^{*}{\Sigma^{*}}{V^{*}}^{T} 
\end{aligned}
\text{, } \widehat{\alpha} = U\Sigma^{-1/2}{\widehat{\alpha^{*}}}  
\end{equation*}
For our algorithm, we put $\alpha=\alpha_m$, $\beta=K^2\beta_m$ and $Q=K$ in \textit{Corollary 1}. This completes the solution for the second stage of the iterative algorithm.\\

\textbf{SKPCA ALGORITHM:}
\\
1. Initialize $\alpha_m$ with the $m$ eigen vectors of $K$ corresponding to the $m$ largest eigen values.
\\2. First for fixed $\alpha_m$, solve the elastic net problem in (11) for $\beta_{m}$.
\\3. Then with the $\beta_m$ obtained in the previous step, update $\alpha_m$ as 
${\alpha_m} = U\Sigma^{-1/2}U^{*}{V^{*}}^{T}$ where $\Sigma^{-1/2}U^{T}K^2{\beta_m} = U^{*}{\Sigma^{*}}{V^{*}}^{T}$ and $K = U\Sigma U^{T}$.
\\4. Repeat steps 2-3 till convergence.
\subsection{Outlier Detection Algorithm using Sparse KPCA}
We employ the same strategy presented in [2] for distinguishing outliers from inliers as explained earlier, except that here we use the $\beta_m$ obtained from the algorithm above (i.e. the sparse solutions) to construct the $W$ matrix in the reconstruction error term.

\section{Theoretical Justification of Sparse KPCA With RBF Kernel}
Assume that we have a set $S_{m}$ of $m$ data points, $\{x_{1},x_{2},\ldots,x_{m}\}$ drawn in i.i.d. fashion from a continuous probability distribution $P$. Consider the subset $S_{n} = \{x_{1},x_{2},\ldots,x_{n}\}$ of $n < m$ points. The elements of $S_{n}$ obey a particular condition, the details of which shall be presented later. 
We shall prove that the eigenvectors of the covariance matrix (henceforth referred to as $C_{m}$) in the kernel space constructed out of all points in $S_{m}$ can be expressed as a linear combination of just $\{\phi(x_{1}),\phi(x_{2}),\ldots,\phi(x_{n})\}$ with high probability for sufficiently large $m$ and $n$. 
         Once again, we assume that the data in the kernel space is centered. An eigenvector, $v$ of $C_{m}$ can be expressed as $v = \sum_{i=1}^m {\alpha_{i}}{\phi}(x_{i})$. Consider a random point $x'$ sampled from $P$. The projection of $x'$ on $v$ in the kernel space is given as $\langle \phi(x'),v \rangle = \sum_{i=1}^m {\alpha_{i}}k(x_{i},x')$, where $k(x,y)$ is the chosen kernel function. For our case, we choose the RBF kernel and thus $k(x,y) = \exp(-\frac{\|x-y\|^2}{2\sigma ^2})$. Without loss of generality, assume $\sigma = 1/\sqrt{2}$ for simplicity. Thus, we have $\langle \phi(x),v \rangle = \sum_{i=1}^n {\alpha_{i}}\exp(-\|x_{i} - x'\|^2) + \sum_{i=n+1}^m {\alpha_{i}}\exp(-\|x_{i} - x'\|^2)$. For $i>n$, define $f(i) = \underset{1 \leq j \leq n}\argmin{ \|x_{i} - x_{j}\|}$. We shall show that $\exp(-\|x_{i} - x'\|^2) \approx \exp(-\|x_{f(i)} - x'\|^2)$ for $i>n$ holds with high probability. 
For $x'$, define $g(x') = \underset{1 \leq j \leq n}\argmin{ \|x_{j} - x'\|}$. Then, for $i>n$ : $\exp(-\|x_{i} - x'\|^2) = \exp(-\|(x_{i} - x_{f(i)}) +(x_{f(i)} - x') \|^2) = \exp(-\|x_{f(i)} - x'\|^2)\times  \exp(-\|x_{i} - x_{f(i)}\|^2) \times \exp(-2\langle (x_{i} - x_{f(i)}),(x_{f(i)} - x') \rangle)$.
         
The third term in the above factorization can further be written as: $\exp(-2\langle (x_{i} - x_{f(i)}),(x_{f(i)} - x') \rangle) = \exp(-2\langle (x_{i} - x_{f(i)}),(x_{f(i)} - x_{g(x')} + x_{g(x')} - x') \rangle) = \exp(-2\langle (x_{i} - x_{f(i)}),(x_{f(i)} - x_{g(x')}\rangle) \times \exp(-2\langle (x_{i} - x_{f(i)}),(x_{g(x')} - x')\rangle)$. 

Finally, we get : $\exp(-\|x_{i} - x'\|^2) = \exp(-\|x_{f(i)} - x'\|^2)\times \exp(-\|x_{i} - x_{f(i)}\|^2) \times \exp(-2\langle (x_{i} - x_{f(i)}),(x_{f(i)} - x_{g(x')}\rangle) \times \exp(-2\langle (x_{i} - x_{f(i)}),(x_{g(x')} - x')\rangle)$.

We will now show that the second, third and fourth terms in the above factorization are individually close to 1, due to which their product is also close to 1 and as a result $\exp(-\|x_{i} - x'\|^2) \approx \exp(-\|x_{f(i)} - x'\|^2)$ with high probability. First we shall show that the second term is close to 1 with high probability. Let $\Pr(\|u - v\| \geq d) = p(d)$ where $u$ \& $v$ are sampled in i.i.d. fashion from the distribution $P$. 
Here, we assume that $p(d) < 1$ for $d>0$. 
From elementary probability, $\Pr(\|x_{i} - x_{j}\| \leq d \text{ for some } j, 1 \leq j \leq n, i > n) = 1- (p(d))^n$ which tends to $1$ for large enough $n$. Thus, $\|x_{i} - x_{f(i)}\| \leq d$ holds with high probability, for sufficiently large $n$. 
For $d = 0.1\sigma = 0.1/\sqrt{2}$, $\exp(-\|x_{i} - x_{f(i)}\|^2) \geq 0.995$. 
Hence, the second term is close to 1 with high probability.	

         We shall deal with the fourth term before we analyze the third term. By the Cauchy-Schwarz inequality, $|\langle (x_{i} - x_{f(i)}),(x_{g(x')} - x')\rangle| \leq \|x_{i} - x_{f(i)}\|.\|x_{g(x')} - x'\|$. We showed in the analysis of the second term that $\|x_{i} - x_{f(i)}\| \leq d$ with high probability. We employ the same analysis for $x'$ instead of $x_{i}$ (and $x_{g(x')}$ instead of $x_{f(i)}$) to conclude that $\|x_{g(x')} - x'\| \leq d$ with high probability. Thus, with $d = 0.1/\sqrt{2}$,  $0.99 \leq \exp(-2\langle (x_{i} - x_{f(i)}),(x_{g(x')} - x')\rangle) \leq 1.01$. Thus, the fourth term is also close to $1$ with high probability.
         
         Finally, we come to the analysis of the third term. Here, $|\langle (x_{i} - x_{f(i)}),(x_{f(i)} - x_{g(x')})\rangle| \leq \|x_{i} - x_{f(i)}\|.\|x_{f(i)} - x_{g(x')}\| \leq d\|x_{f(i)} - x_{g(x')}\|$ from the analysis of the second term. Let $S$ denote a subset of any $n$ distinct points out of the $m$ points in $S_{m}$. There are ${m}\choose{n}$ such subsets. Define $d_{max}^S = \underset{i \in S,j \in S}{\text{max }}
\mathrm \|x_{i}-x_{j}\|$. With this definition, $|\langle (x_{i} - x_{f(i)}),(x_{f(i)} - x_{g(x')})\rangle| \leq d.d_{max}^{S_{n}}$. Let $\Pr(d_{max}^S \geq \tilde{d} \text{ for a particular } S) = \tilde{p}(\tilde{d}) = 1-(1-p(\tilde{d}))^{{n}\choose{2}}$. Here again, we have $\tilde{p}(\tilde{d}) < 1$ for $\tilde{d} > 0$ as $p(\tilde{d}) < 1$ for $\tilde{d} > 0$. Thus, $\Pr(d_{max}^S \leq \tilde{d} \text{ for some } S \text{ out of all possible choices}) = 1- (\tilde{p}(\tilde{d}))^{{m}\choose{n}}$ which also tends to $1$ for $m >> n$. Without loss of generality, assume that $S_{n}$ satisfies the condition $d_{max}^{S_{n}} \leq \tilde{d}$. This is the exact condition on the elements of $S_{n}$ which was mentioned in the beginning of this section. Taking $\tilde{d} = 0.3/\sqrt{2}$,  $0.97 \leq \exp(-2\langle (x_{i} - x_{f(i)}),(x_{f(i)} - x_{g(x')})\rangle) \leq 1.03$. Thus, even the third term is also close to $1$ with high probability.

Thus, $\exp(-\|x_{i} - x'\|^2) \approx \exp(-\|x_{f(i)} - x'\|^2)$ holds with high probability. So, we have $\langle \phi(x),v \rangle \approx \sum_{i=1}^n {\alpha_{i}}\exp(-\|x_{i} - x'\|^2) + \sum_{i=n+1}^m {\alpha_{i}}\exp(-\|x_{f(i)} - x'\|^2)$ with high probability. This can be rewritten as $\langle \phi(x),v \rangle \approx \sum_{i=1}^n {\beta_{i}}\exp(-\|x_{i} - x'\|^2)$, where $\beta_{i} = (\alpha_{i}+\sum_{j>n \mid f(j)=i} {\alpha_{j}}$). Therefore, $v = \sum_{i=1}^m {\alpha_{i}}{\phi}(x_{i}) \approx \sum_{i=1}^n {\beta_{i}}{\phi}(x_{i})$ with high probability, where $\beta_{i}$'s are as defined before. This finishes our proof.

Note that one can design a naive algorithm based on the above proof, but its combinatorial nature would render it highly inefficient. We conjecture that our algorithm (or any other sparse KPCA algorithm) does this subset selection task efficiently and perhaps approximately by solving a convex problem instead.
         
\section{Experiments}

We have compared the performance of our SKPCA algorithm with standard KPCA, naive thresholding of the $m$ largest magnitude components of the KPCA coefficients (with $m$ = average number of non-zero components per PC obtained through SKPCA) as well the algorithm given in [3]. 
We have compared our method with naive thresholding because it has been noted in [1] that naive thresholding works almost as well as their sparse PCA method.
We have tested these methods on 5 real world datasets - MNIST (Fig. 1), Fashion MNIST[12] (Fig. 2), Satimage2[13] (Fig. 3), ETH-80 dataset (Fig. 4) and Internet Advertisements dataset[14] (Fig. 5). 
Our experiments reveal that our SKPCA algorithm provides high sparsity without compromising on accuracy. The accuracy using our SKPCA algorithm is nearly the same as KPCA, which is not possible with naive thresholding at low sparsity and the sparsity obtained is in general much more than that obtained from the algorithm in [3], for the same accuracy. 
Also, our algorithm has just 1 sparsity controlling parameter which is the L1-ratio = $\lambda_j/\lambda$ $\forall j$, whereas the algorithm in [3] has 5 parameters - $\rho,\lambda,\lambda_{1,k},\epsilon^{abs},\epsilon^{rel}$.
The metric used to measure accuracy in our experiments is F1-score (standard for outlier detection tasks).
We obtained 3 plots in our simulations. The first one is a box plot to show the variability of F1-score over different training and test data subsets (we performed the same experiment 10 times on randomly chosen training and test data subsets out of the entire dataset). However, we did not show the variability of the algorithm in [3] for reasons explained later.
The second one is a F1-score vs. sparsity (measured as the percentage of non-zero coefficients in the kernel space PCs) curve for all the aforementioned methods. The F1-score vs. sparsity curves were constructed by trying out the various methods on a range of sparsity controlling parameters  for a randomly chosen subset of training and test data. The KPCA horizontal line in the F1-score vs. sparsity curves is provided just for reference and it has all non-zero coefficients. The third one contains ROC curves for all the aforementioned methods.
The number of PCs used for MNIST, Fashion MNIST, ETH-80, Satimage2 and Internet Ads were 15, 15, 15, 7 and 24 respectively. We took $\sigma^{2}$ in the RBF kernel to be the average of $\|x_i-x_j\|^2$ for all pairs $i,j$ such that $i \neq j$.  
\justify
\textbf{MNIST:}
Similar to the experiment performed in [2], we chose 0 as the inlier class and all digits from 1-9 as the outlier class. We used a training set of 3000 
inliers and a test set consisting of 3000 inliers and 3000 outliers. The variability plot is for L1-ratio=0.7. The mean and standard deviation of sparsity for the 10 trials were 3.35\% and 0.13\% respectively. The ROC curve of our algorithm is also for L1-ratio=0.7, while the ROC curve of the algorithm in [3] is for $\rho = 0.02, \lambda=0.001,\lambda_{1,k}=0.01,\epsilon^{abs}=0.01,\epsilon^{rel}=0.0001$. 
The area under ROC curve (AUROC) values for our SKPCA algorithm, standard KPCA, the algorithm in [3] and naive thresholding were 0.974, 0.986, 0.962 and 0.957 respectively.

\justify
\textbf{Fashion MNIST:}
This is a recent dataset which is very similar to MNIST but much more challenging. Here, we took item 0 (t-shirts and some other tops) as the inlier class and item 1 (pants) as the outlier class. Training and test dataset size was the same as in MNIST. The variability plot is for L1-ratio=0.3. The mean and standard deviation of sparsity for the 10 trials were 8.43\% and 0.24\% respectively. The ROC curve of our algorithm is also for L1-ratio=0.3, while the ROC curve of the algorithm in [3] is for $\rho = 0.05, \lambda=0.002,\lambda_{1,k}=0.006,\epsilon^{abs}=0.01,\epsilon^{rel}=0.0001$.
The AUROC values for our SKPCA algorithm, standard KPCA, the algorithm in [3] and naive thresholding were 0.919, 0.898, 0.872 and 0.877 respectively.

\justify
\textbf{Satimage2:}
The data was already presented as inliers and outliers. Training set size was 400 inliers and test set consisted of 500 inliers and 71 outliers (only that many were available). The variability plot is for L1-ratio=0.6. 
The mean and standard deviation of sparsity for the 10 trials were 5.55\% and 0.33\% respectively. The ROC curve of our algorithm is also for L1-ratio=0.6, while the ROC curve of the algorithm in [3] is for $\rho = 0.07, \lambda=0.003,\lambda_{1,k}=0.016,\epsilon^{abs}=0.01,\epsilon^{rel}=0.0001$.
The AUROC values for our SKPCA algorithm, standard KPCA, the algorithm in [3] and naive thresholding were 0.963, 0.958, 0.942 and 0.935 respectively.

\justify
\textbf{ETH-80:}
We used the apples in this data set as inliers and tomatoes as outliers (they have nearly the same color and shape making it more challenging). Training set size was 300 inliers and test set consisted of 111 inliers and 111 outliers. The variability plot and is for L1-ratio=0.3. The mean and standard deviation of sparsity for the 10 trials were 11.07\% and 0.70\% respectively. The ROC curve of our algorithm is also for L1-ratio=0.3, while the ROC curve of the algorithm in [3] is for $\rho = 0.1, \lambda=0.006,\lambda_{1,k}=0.012,\epsilon^{abs}=0.01,\epsilon^{rel}=0.0001$.
The AUROC values for our SKPCA algorithm, standard KPCA, the algorithm in [3] and naive thresholding were 0.859, 0.853, 0.841 and 0.836 respectively.

\justify
\textbf{Internet Ads:}
The data was already presented as inliers and outliers. Training set size was 600 inliers and test set consisted of 380 inliers and 380 outliers. The variability plot and is for L1-ratio=0.4. The mean and standard deviation of sparsity for the 10 trials were 2.60\% and 0.29\% respectively. The ROC curve of our algorithm is also for L1-ratio=0.4, while the ROC curve of the algorithm in [3] is for $\rho = 0.008, \lambda=0.009,\lambda_{1,k}=0.019,\epsilon^{abs}=0.01,\epsilon^{rel}=0.0001$.
The AUROC values for our SKPCA algorithm, standard KPCA, the algorithm in [3] and naive thresholding were 0.783, 0.785, 0.744 and 0.739 respectively.\\

\begin{figure}[!h]
\centering 
\subfloat[F1-Score Variability plot for MNIST]{
	\includegraphics[width=36mm, height=39mm]{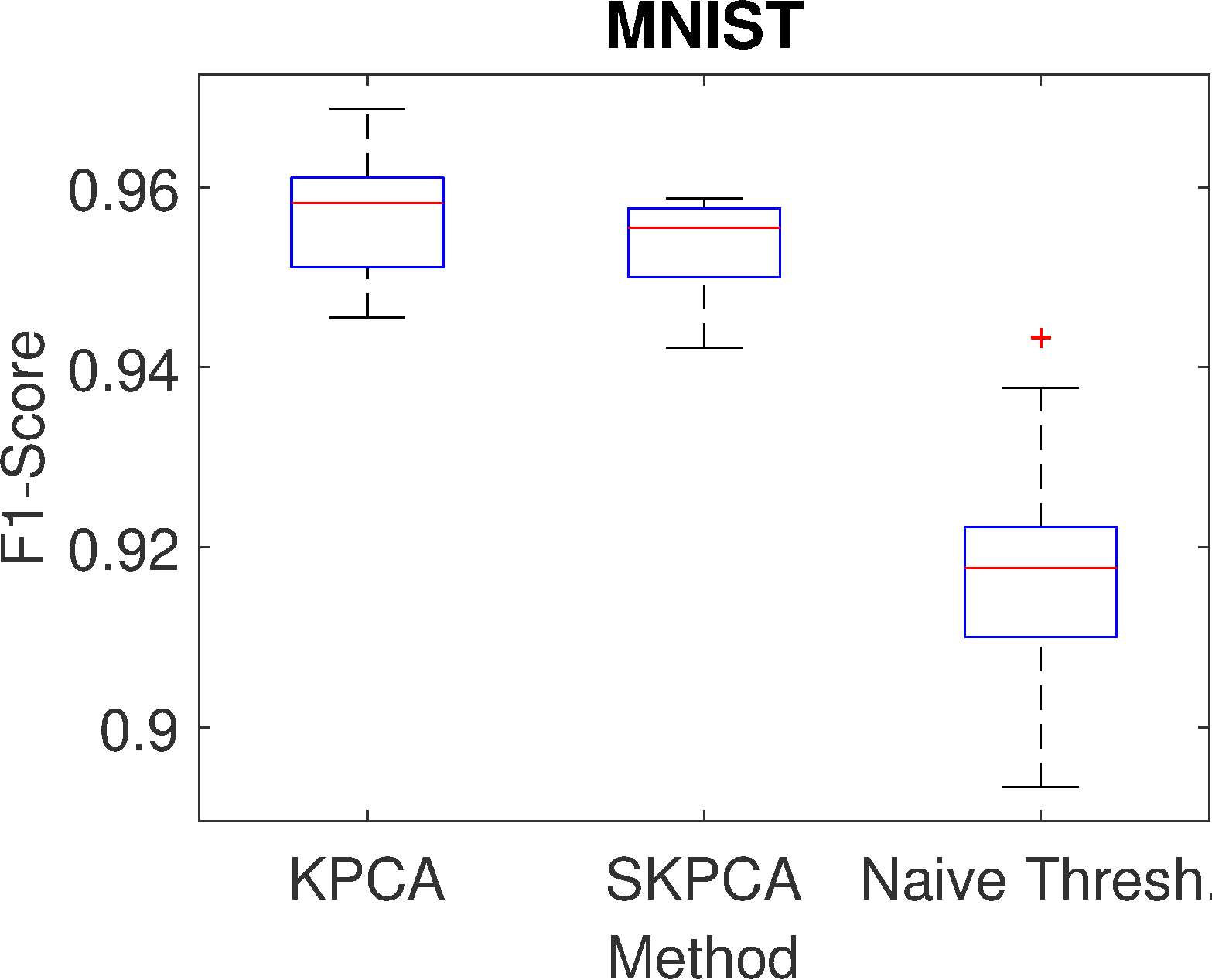} } 
\hspace{0.3 in}   
\subfloat[F1-Score vs. Sparsity plot for MNIST]{
	\includegraphics[width=36mm, height=39mm]{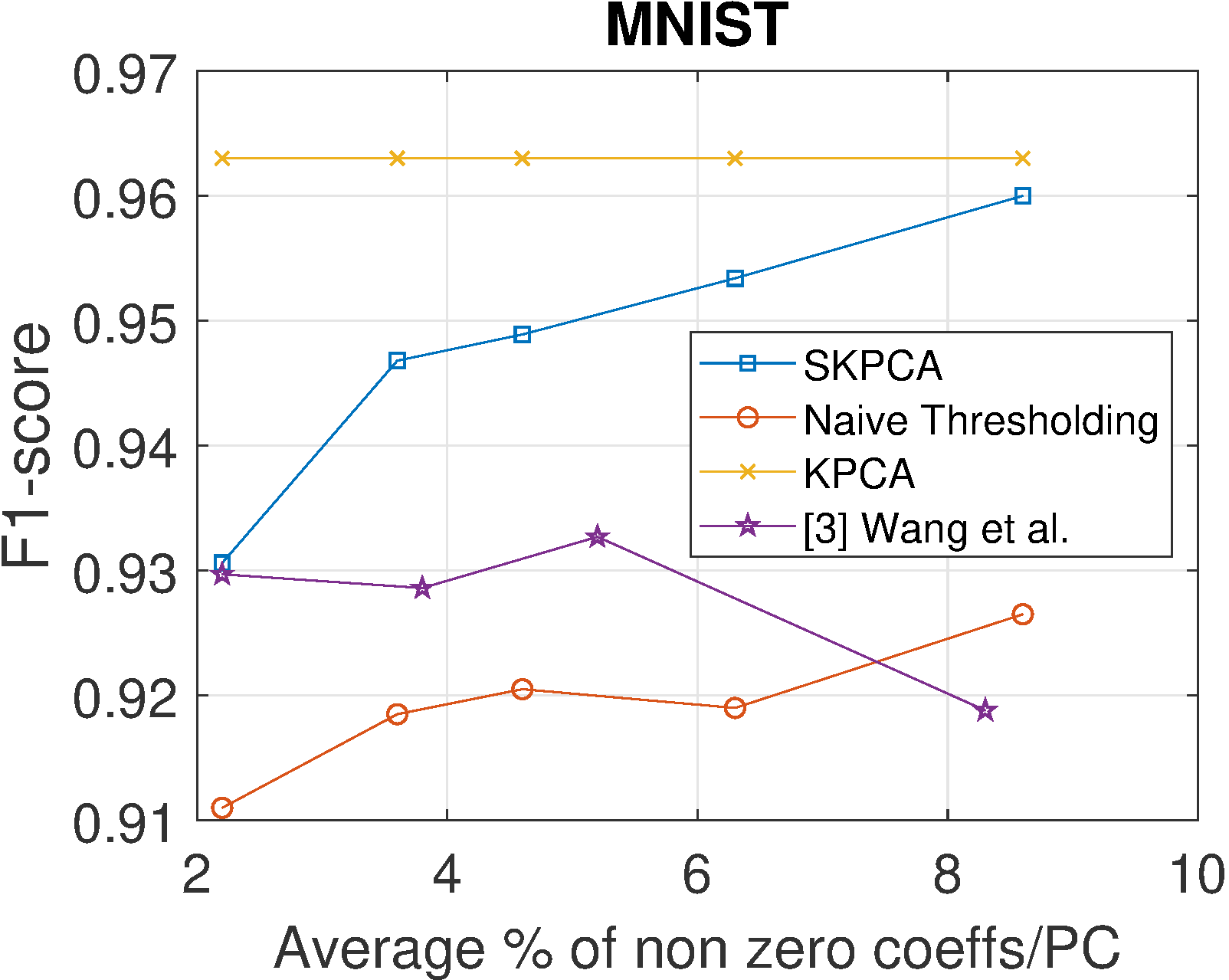}}  
\hspace{0.2 in} 
\subfloat[ROC curves for MNIST]{
	\includegraphics[width=36mm, height=39mm]{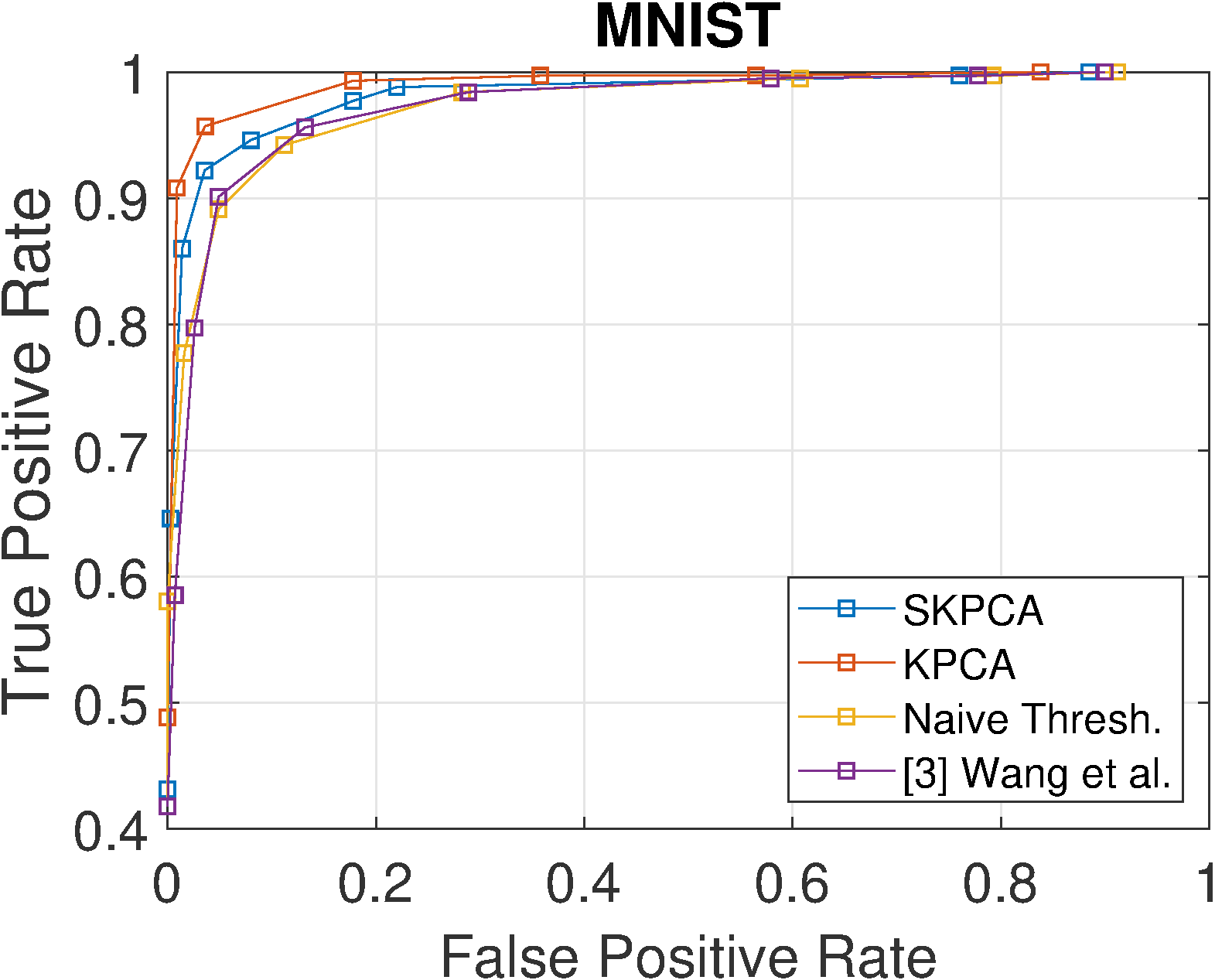}}  
\caption{Plots for MNIST}
\end{figure}
\begin{figure}[!h]
\centering 
\subfloat[F1 score Variability plot for Fashion MNIST]{
	\includegraphics[width=36mm, height=38mm]{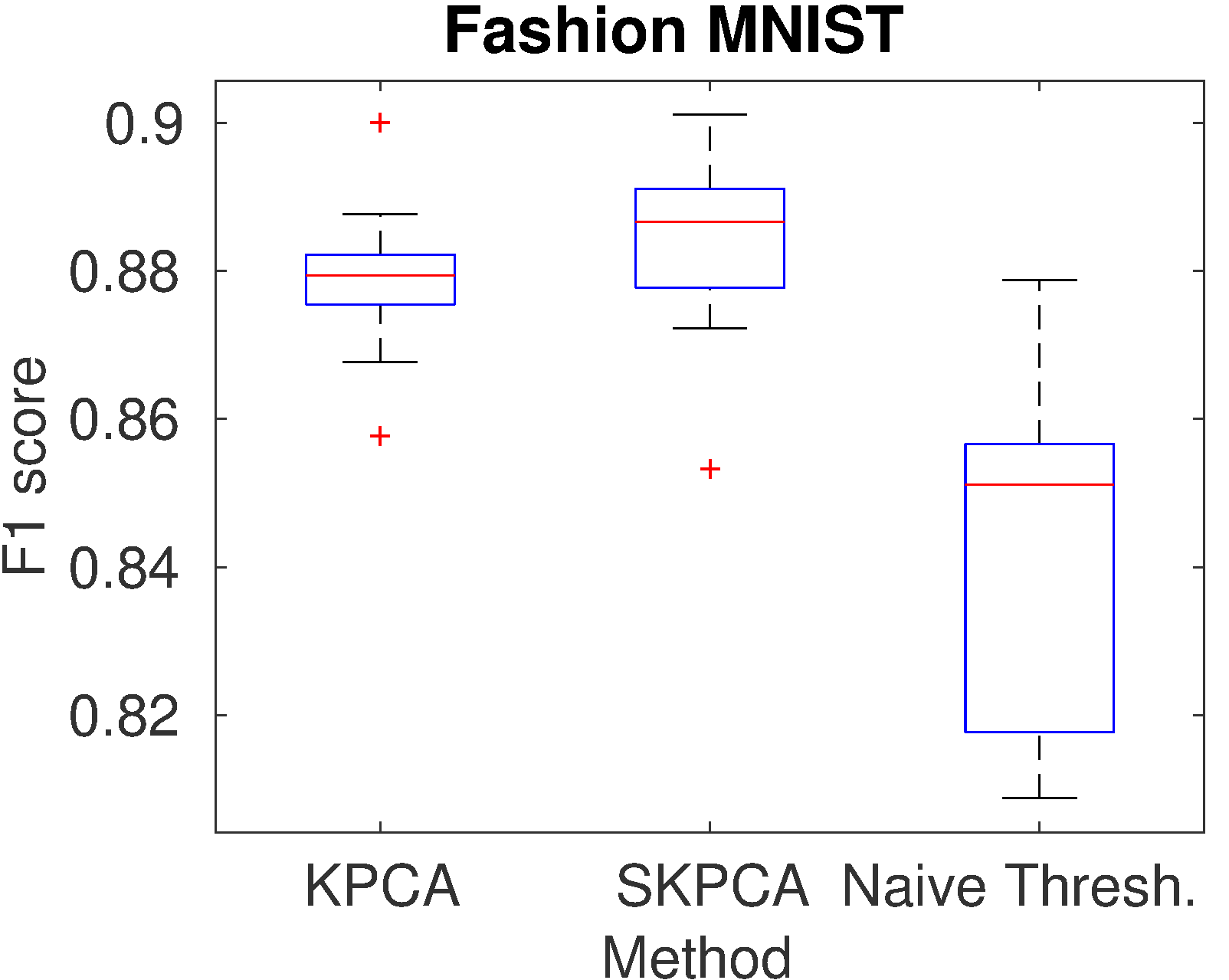} }
\hspace{0.3 in}
\subfloat[F1 score vs. Sparsity plot for Fashion MNIST]{
	\includegraphics[width=36mm, height=38mm]{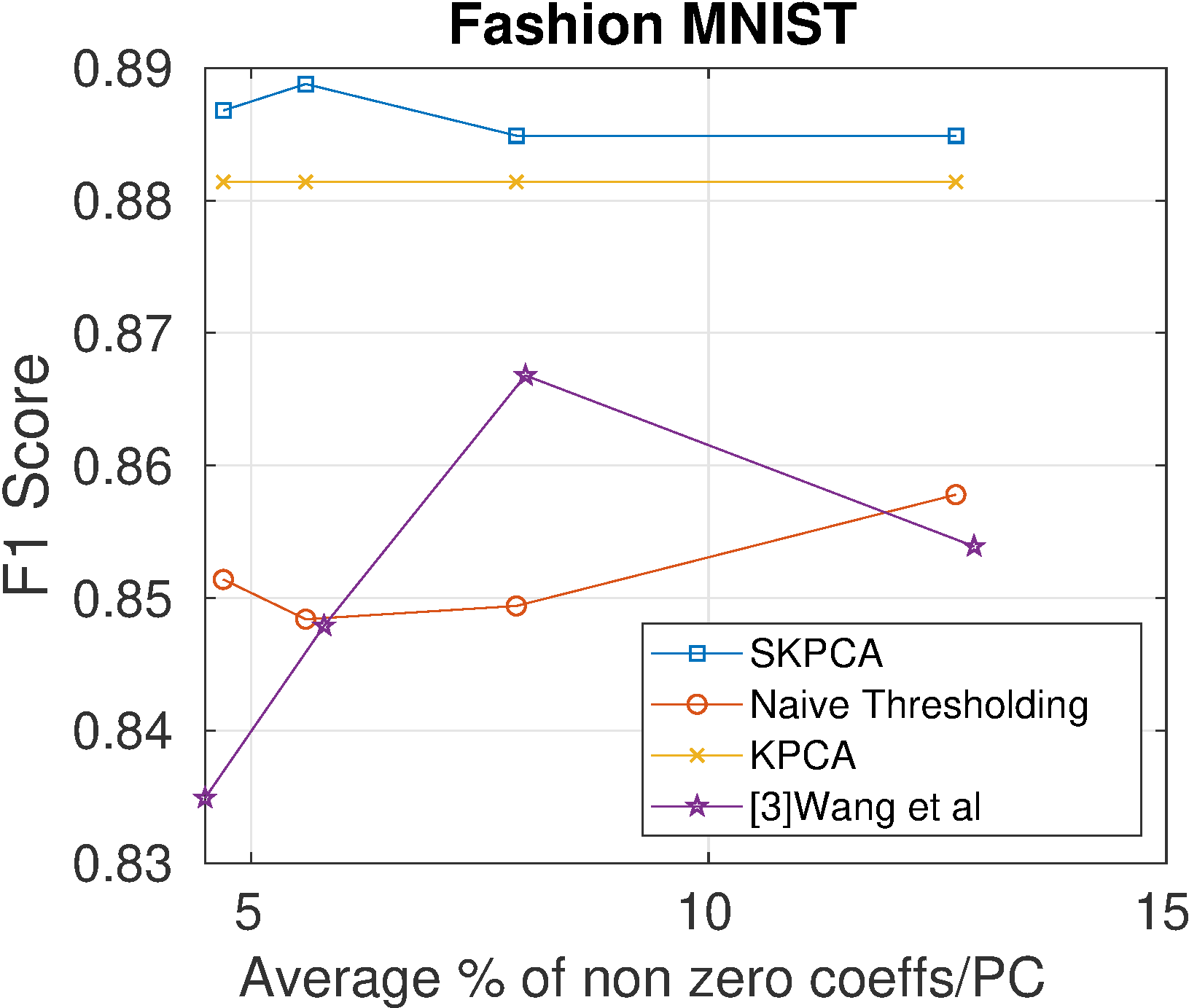}}  
\hspace{0.2 in} 
\subfloat[ROC curves for Fashion MNIST]{
	\includegraphics[width=36mm, height=39mm]{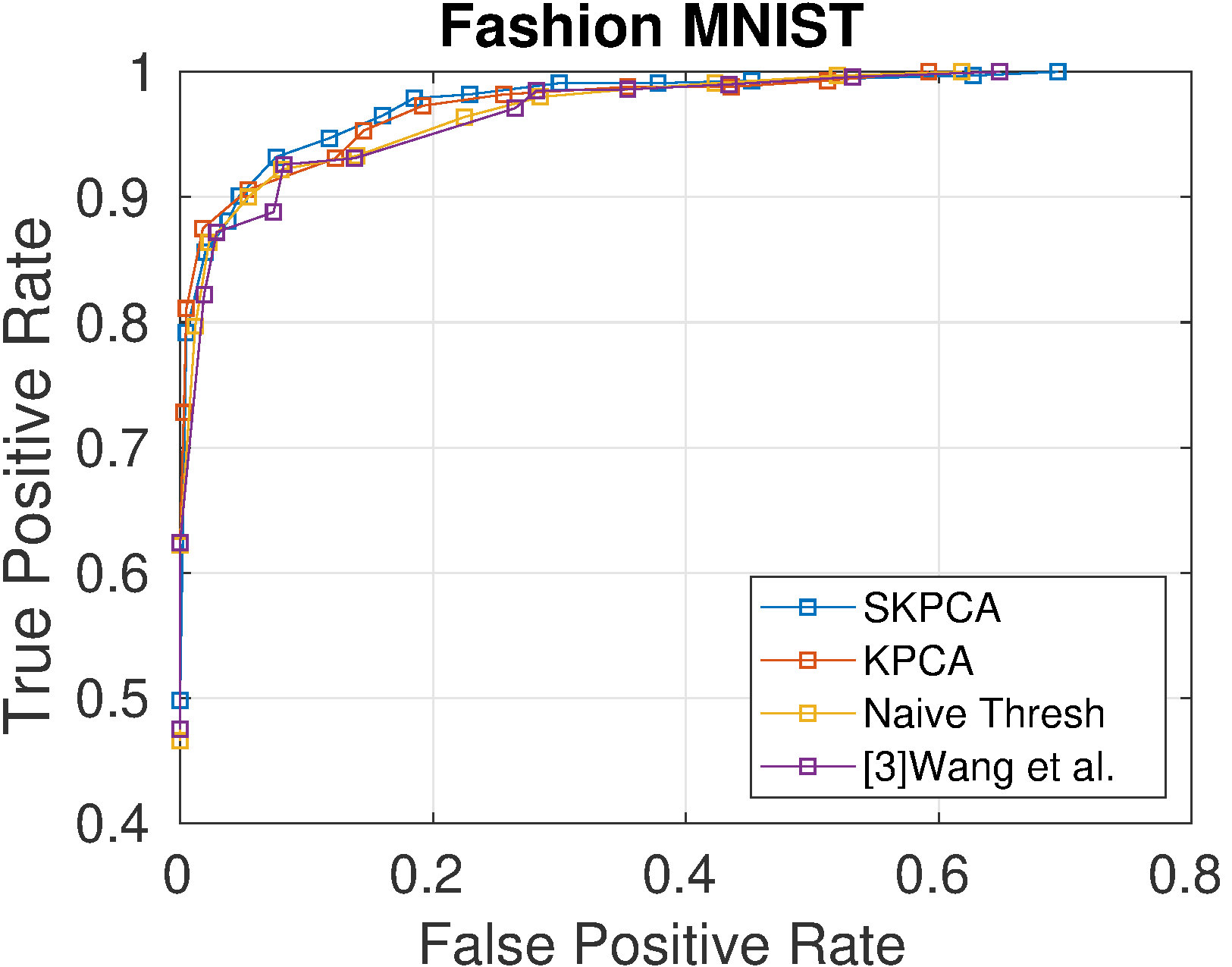}}  
\caption{Plots for Fashion MNIST}
\end{figure}
\begin{figure}[!h]
\centering 
\subfloat[F1-Score Variability plot for Satimage2]{
	\includegraphics[width=36mm, height=38mm]{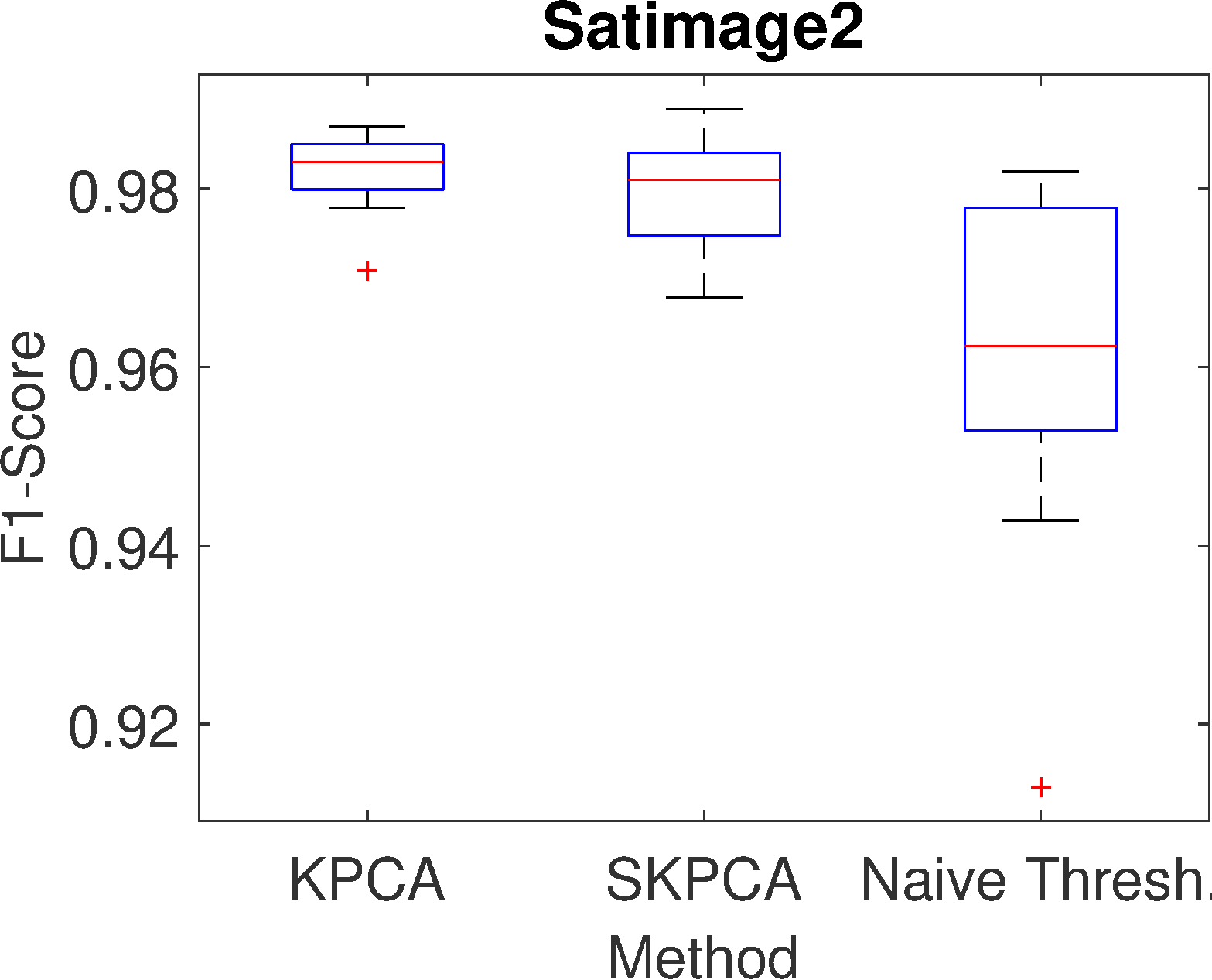} } 
\hspace{0.3 in}  
\subfloat[F1-Score vs. Sparsity plot for Satimage2]{
	\includegraphics[width=36mm, height=38mm]{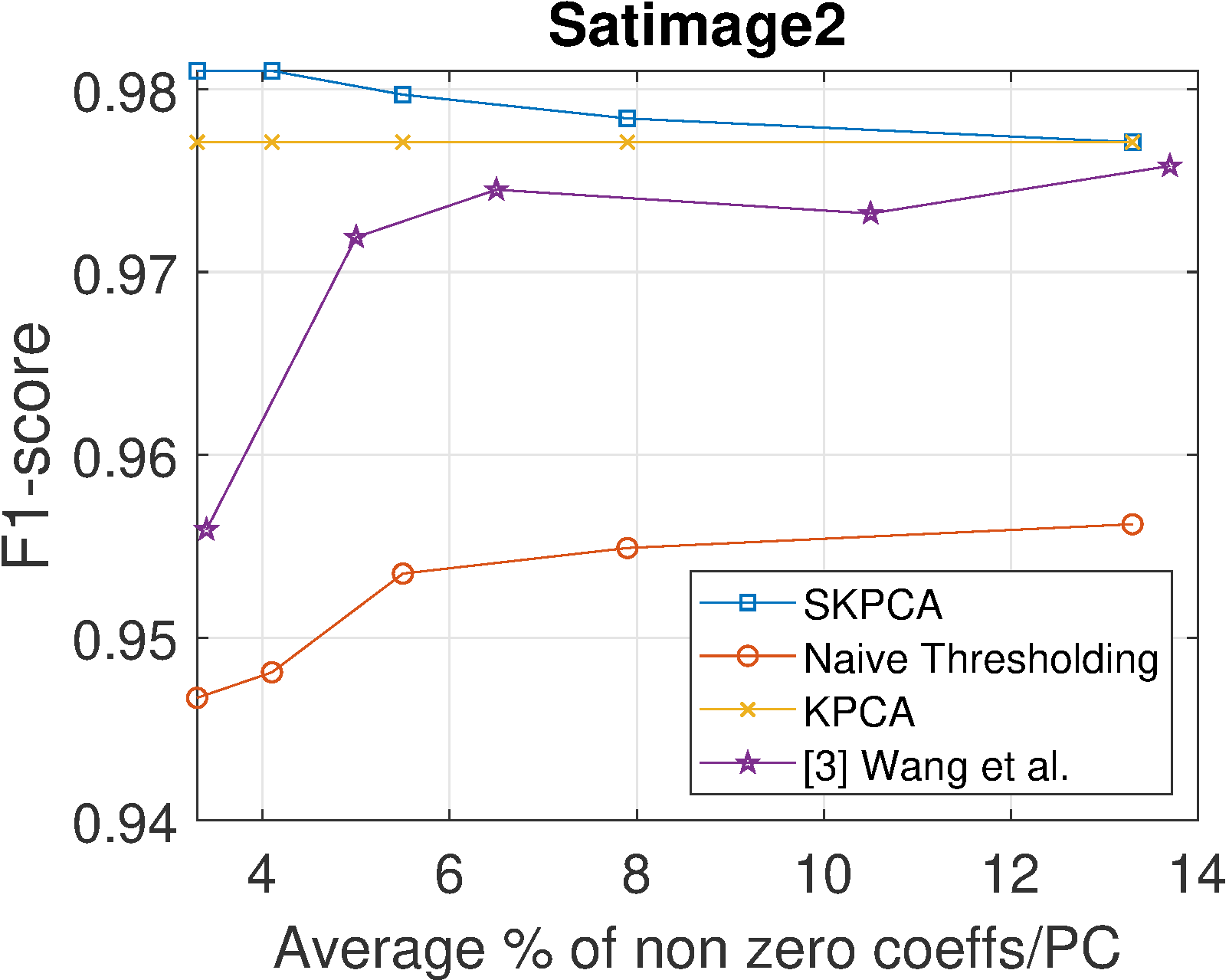}}
\hspace{0.2 in} 
\subfloat[ROC curves for Satimage2]{
	\includegraphics[width=36mm, height=39mm]{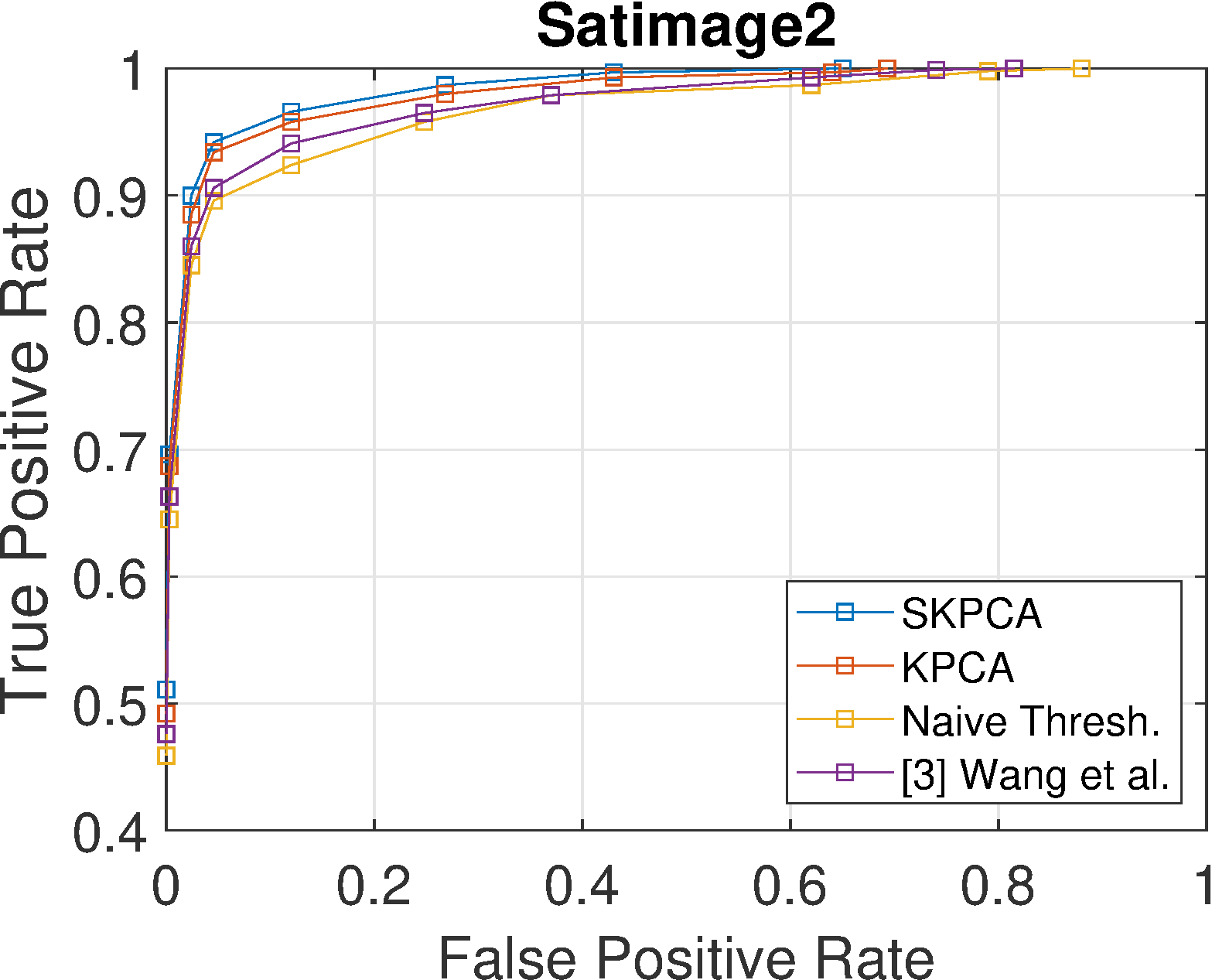}}  
\caption{Plots for Satimage2}
\end{figure} 
\begin{figure}[!h]
\centering  
\subfloat[F1 score Variability plot for ETH-80]{
	\includegraphics[width=36mm, height=38mm]{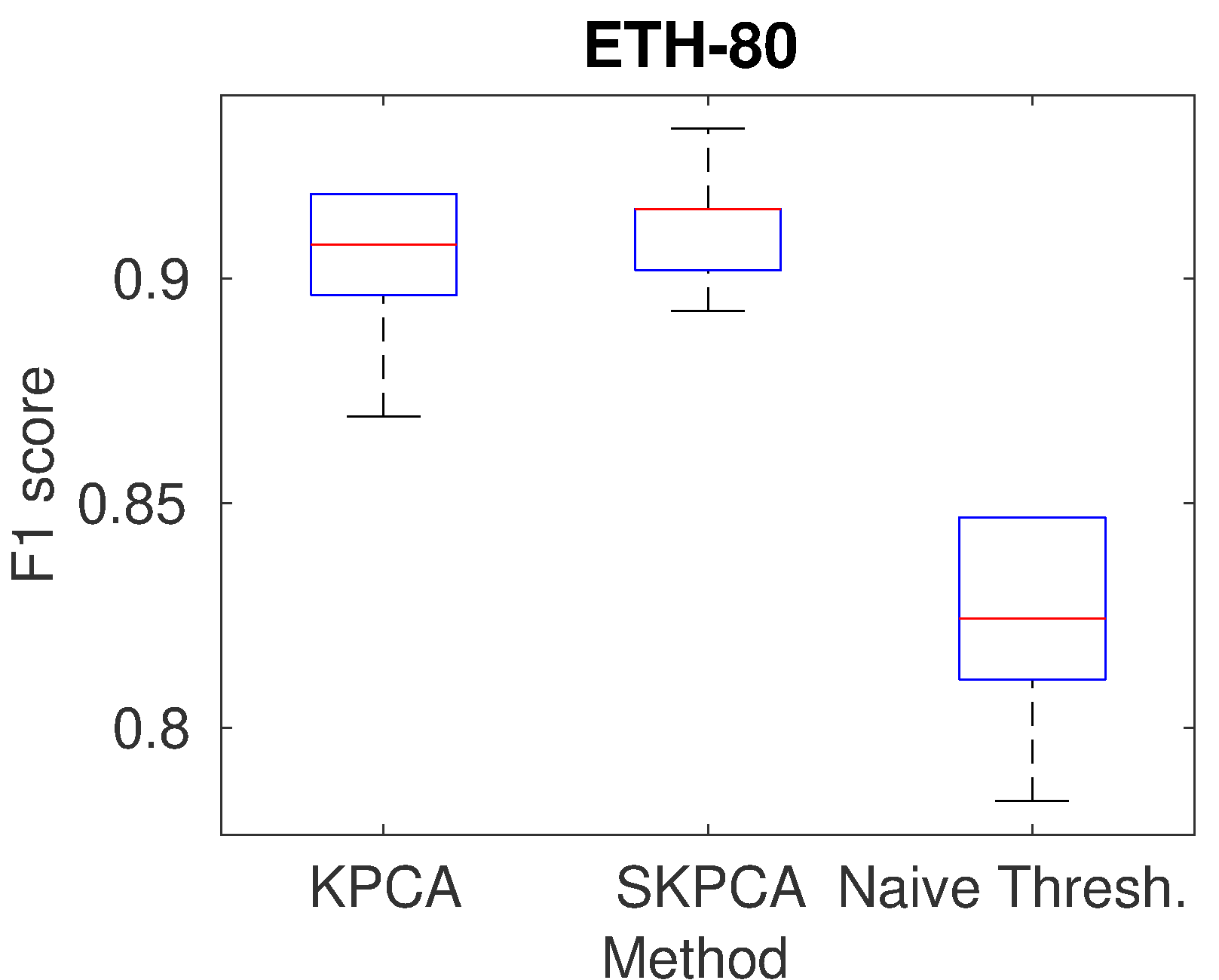} } 
\hspace{0.3 in}
\subfloat[F1 score vs. Sparsity plot for ETH-80]{
	\includegraphics[width=36mm, height=38mm]{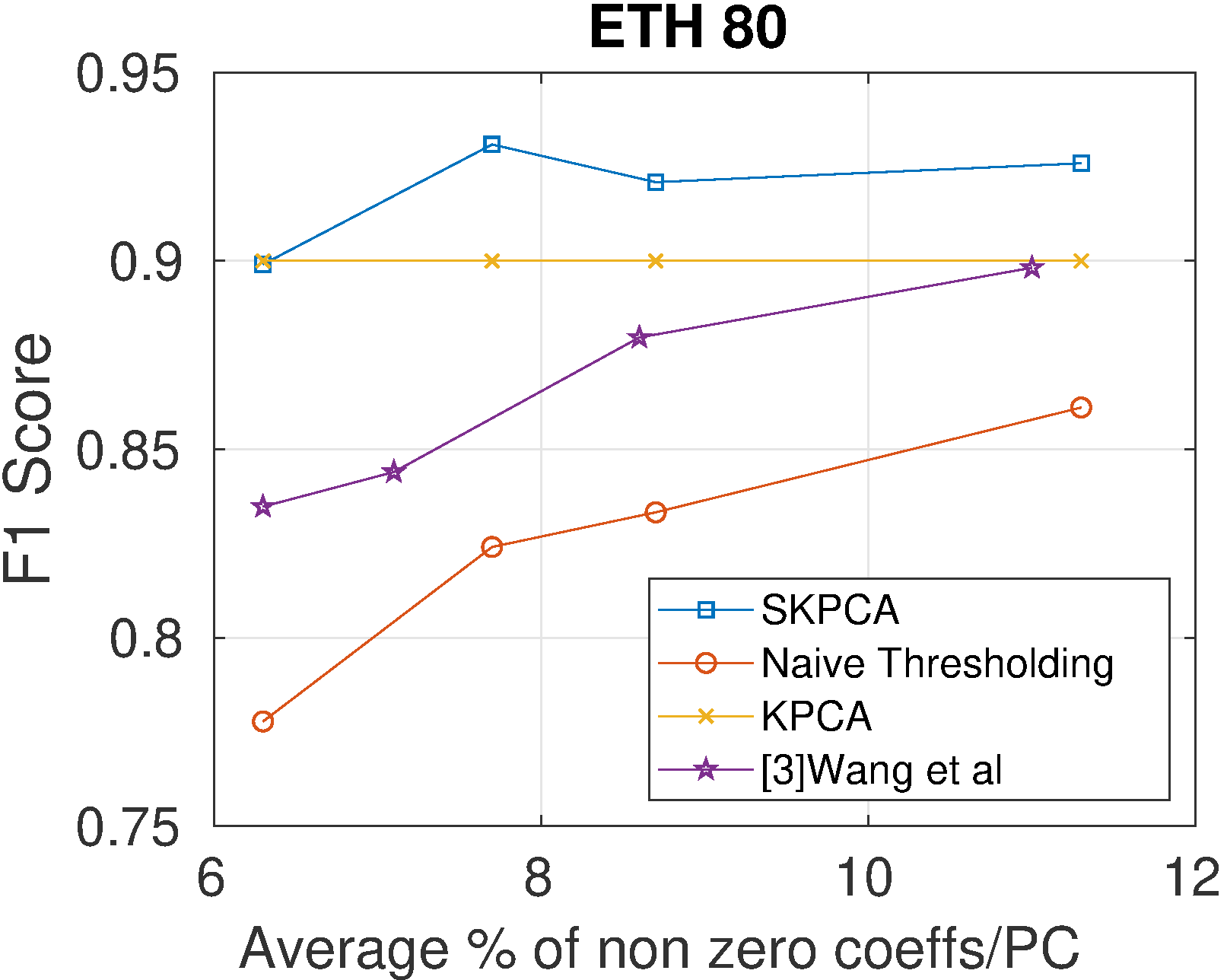}} 
\hspace{0.2 in} 
\subfloat[ROC curves for ETH-80]{
	\includegraphics[width=36mm, height=39mm]{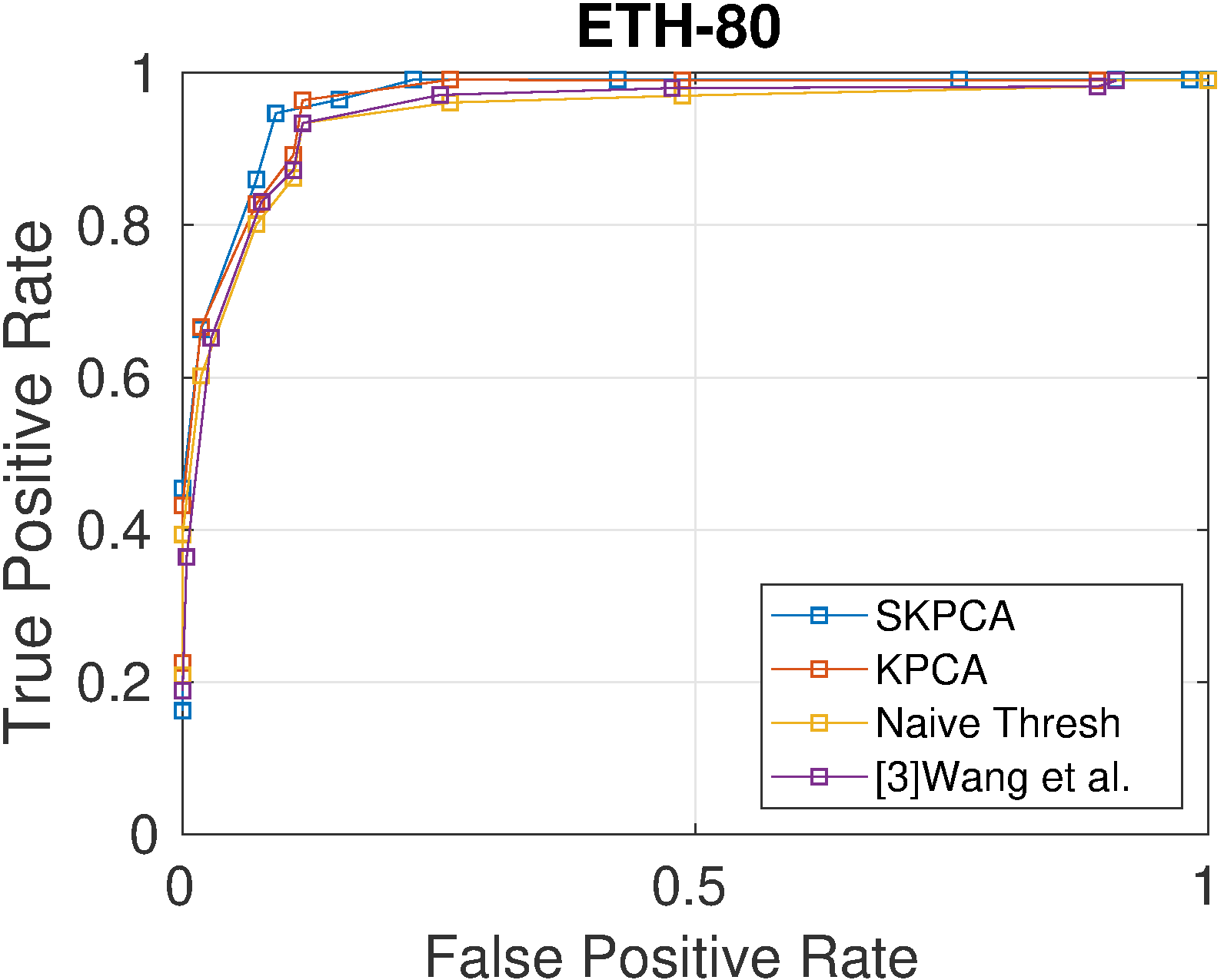}}  
\caption{Plots for ETH-80}
\end{figure}
\begin{figure}[!h]
\centering 
\subfloat[F1-Score Variability plot for Internet Ads]{
	\includegraphics[width=36mm, height=38mm]{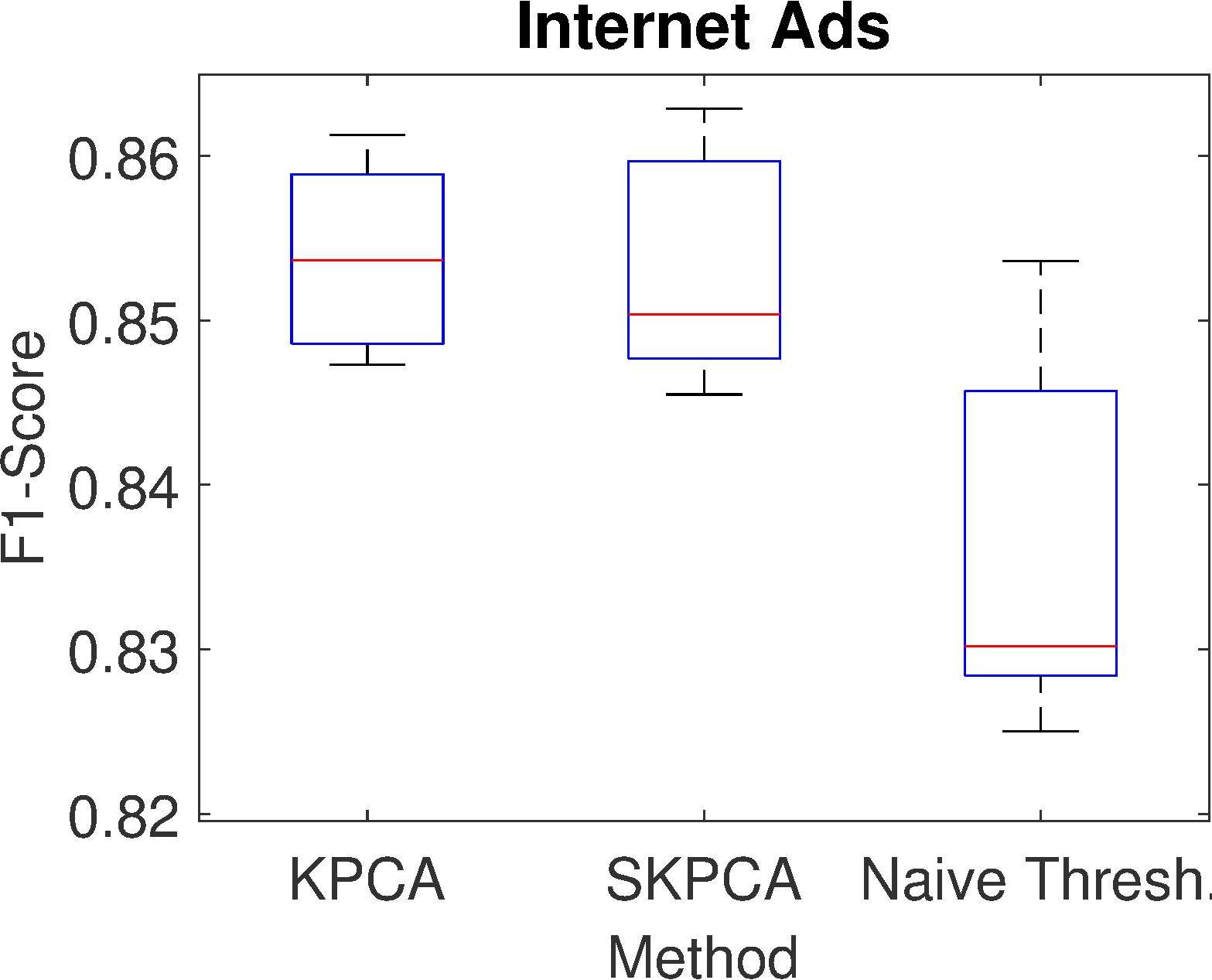} } 
\hspace{0.3 in}  
\subfloat[F1-Score vs. Sparsity plot for Internet Ads]{
	\includegraphics[width=36mm, height=38mm]{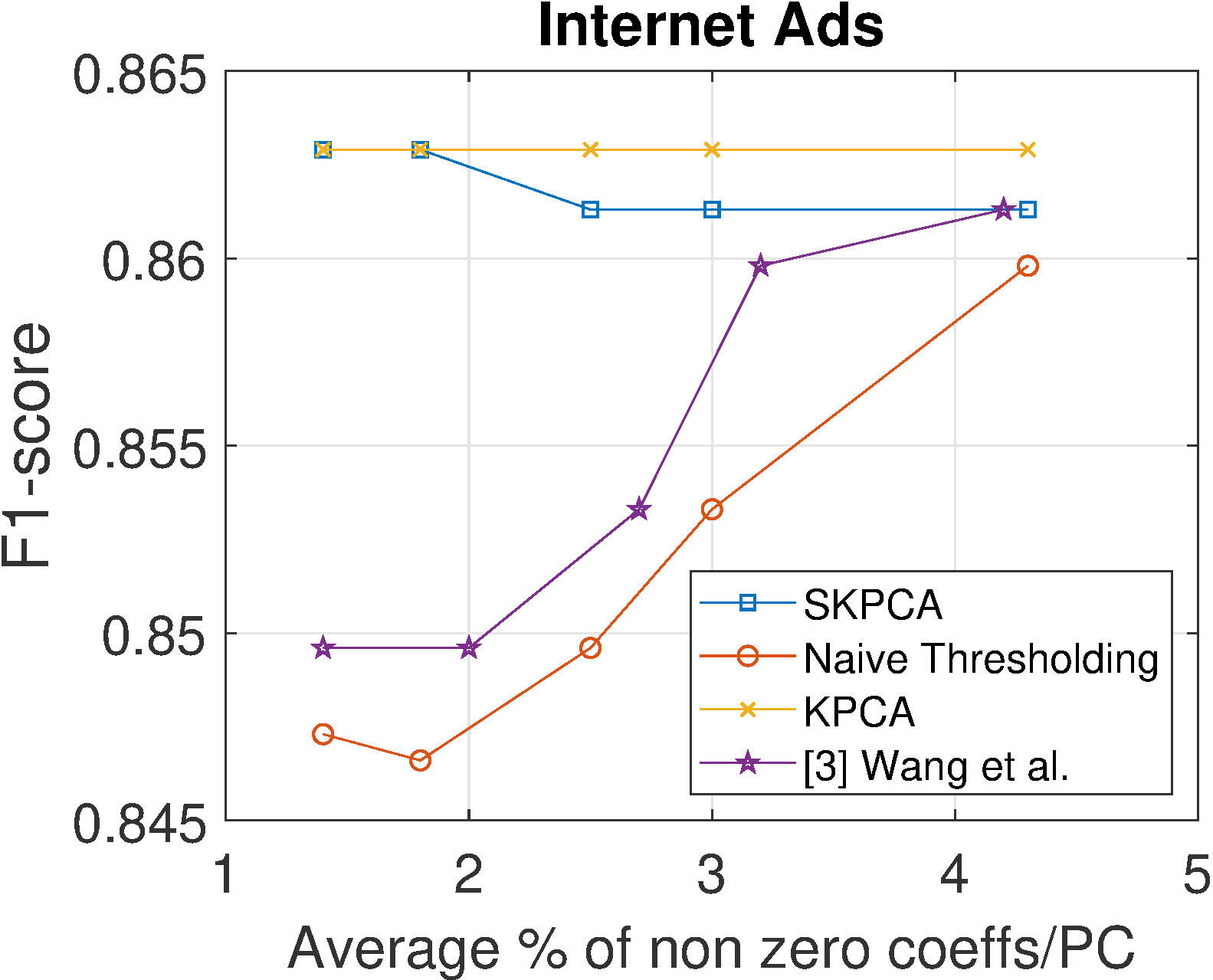} } 
\hspace{0.2 in} 
\subfloat[ROC curves for Internet Ads]{
	\includegraphics[width=36mm, height=39mm]{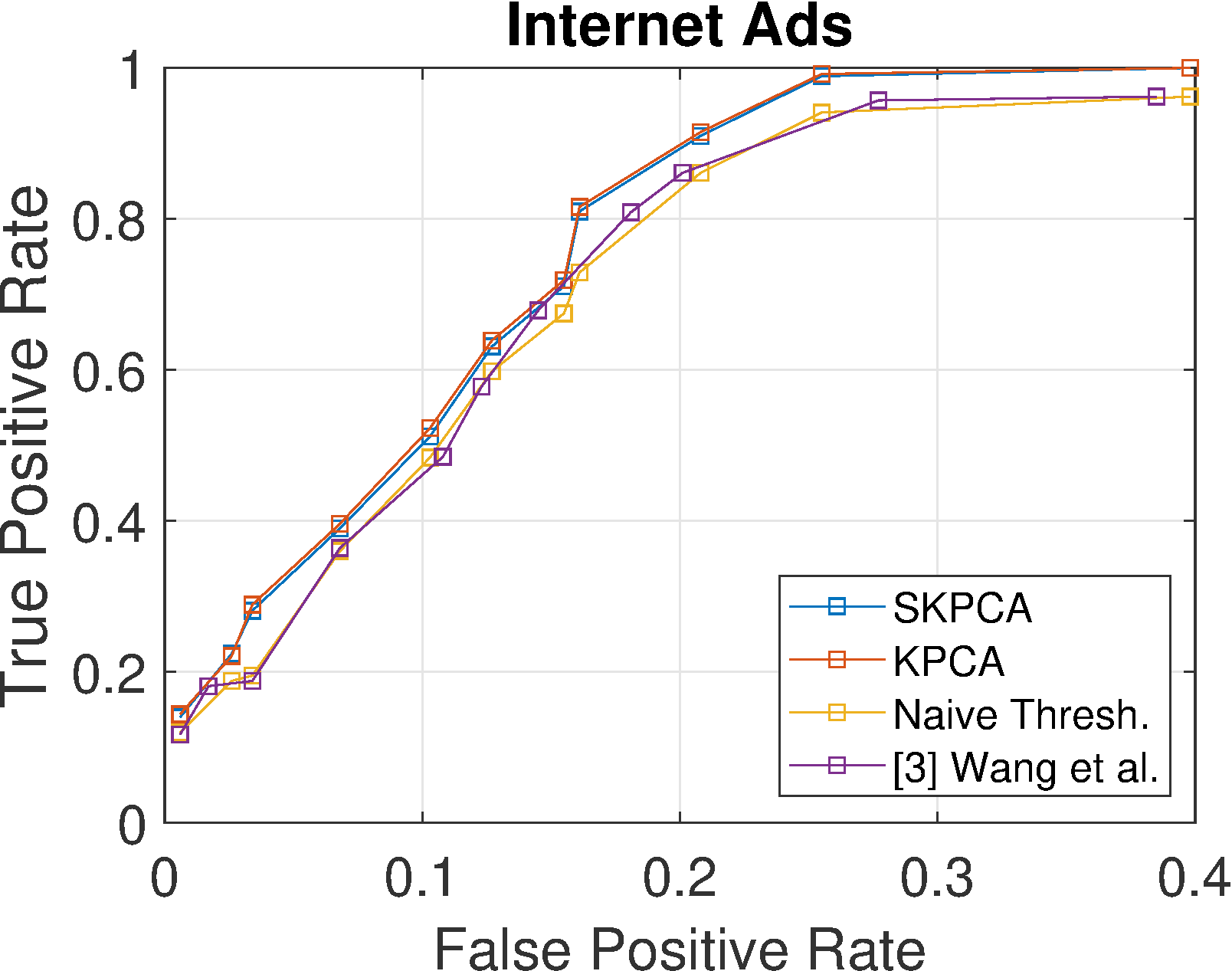}}  
\caption{Plots for Internet Ads}
\end{figure}

It can be seen from the variability plots that the variance of F1-score is lower for SKPCA as compared to naive thresholding for all 5 datasets, which was also mentioned in [1] for sparse PCA. The F1-score vs. sparsity plots show that the F1-score of our SKPCA method is the closest to ordinary KPCA for all 5 datasets (and even better than it for 3 datasets), in comparison to the other 2 methods, over the entire sparsity range in consideration. Also, the AUROC value of our method is more than that of the other 2 methods for all 5 datasets and even more than that of ordinary KPCA for 3 datasets-Fashion MNIST, Satimage2 and ETH-80.

We have not added the variability of the method in [3] because we found out in our experiments that the parameters involved in their algorithm are very sensitive to the chosen data subset (resulting in large differences in F1-score and sparsity for different data subsets). This is not the case with our algorithm, i.e. our algorithm does not require much tuning of parameters over randomly chosen data subsets (for the same problem) as compared to [3], for near optimal performance. Table \Rmnum{1} lists the F1-score and sparsity (i.e. the \% of non-zero coefficients per PC) obtained using our SKPCA algorithm and the method in [3] for the MNIST case over 10 randomly chosen training and test data subsets with fixed parameters for both algorithms - L1-ratio=0.7 for our algorithm and $\rho = 0.02, \lambda=0.001,\lambda_{1,k}=0.01,\epsilon^{abs}=0.01,\epsilon^{rel}=0.0001$ for the algorithm in [3].

\begin{table}[!h]
\centering
\includegraphics[width=0.9\linewidth]{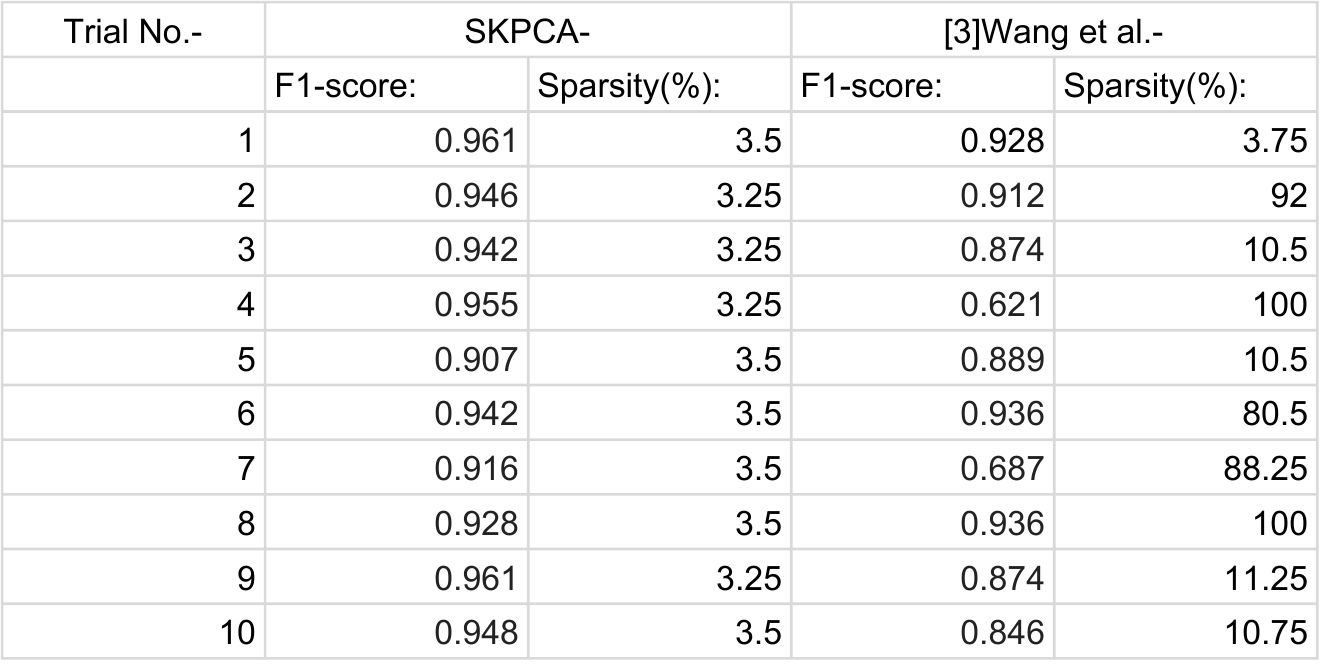}
\caption{F1-score \& sparsity of our SKPCA algorithm and the algorithm in [3] with fixed parameters for MNIST.}
\label{tab:ExcelTable}
\end{table}

\section{CONCLUSION}
In this paper, we presented a novel algorithm for sparse KPCA which outperforms the method in [3] and is comparable to KPCA in terms of accuracy, while providing high sparsity. We also mathematically showed the validity of sparsifying KPCA with the RBF kernel, which is the first attempt in this direction, to the best of our knowledge. 
We showed its successful application for outlier detection on 5 real world data sets. The next step would be to explore the performance of this sparse KPCA algorithm on other applications where KPCA is employed.

\addtolength{\textheight}{-12cm}   






\begin{thebibliography}{150}

\bibitem{c1} Zou, Hui, Trevor Hastie, and Robert Tibshirani. "Sparse principal component analysis." Journal of computational and graphical statistics 15.2 (2006): 265-286.
\bibitem{c2} Hoffmann, Heiko. "Kernel PCA for novelty detection." Pattern Recognition 40.3 (2007): 863-874.
\bibitem{c3} Wang, Duo, and Toshihisa Tanaka. "Sparse kernel principal component analysis based on elastic net regularization." Neural Networks (IJCNN), 2016 International Joint Conference on. IEEE, 2016.
\bibitem{c4} Zou, Hui, and Trevor Hastie. "Regularization and variable selection via the elastic net." Journal of the Royal Statistical Society: Series B (Statistical Methodology) 67.2 (2005): 301-320.
\bibitem{c5} Sch{\"o}lkopf, Bernhard, et al. "Support vector method for novelty detection." Advances in neural information processing systems. 2000.
\bibitem{c6} Tax, David MJ, and Robert PW Duin. "Support vector domain description." Pattern recognition letters 20.11 (1999): 1191-1199.
\bibitem{c7} Sch{\"o}lkopf, Bernhard, Alexander Smola, and Klaus-Robert Müller. "Kernel principal component analysis." International Conference on Artificial Neural Networks. Springer, Berlin, Heidelberg, 1997.
\bibitem{c8} Alexander Smola, O. Mangasarian and Sch{\"o}lkopf, Bernhardand Klaus-Robert Müller. "Sparse kernel feature
analysis," Tech. Rep.,1999
\bibitem{c9} M. E. Tipping, "Sparse kernel principal component analysis," in Advances in Neural Information Processing Systems 13. MIT Press,2001,
pp. 633--539.
\bibitem{c10} M. E. Tipping and C. M. Bishop, "Probabilistic principal component
analysis," Journal of the Royal Statistical Society, Series B, vol. 61,pp.
611-522,1999.


\bibitem{c11}Achlioptas, Dimitris, Frank McSherry, and Bernhard Schölkopf. "Sampling techniques for kernel methods." Advances in neural information processing systems. 2002.


\bibitem{c12}
Xiao, Han, Kashif Rasul, and Roland Vollgraf. "Fashion-mnist: a novel image dataset for benchmarking machine learning algorithms." arXiv preprint arXiv:1708.07747 (2017).
\bibitem{c13}
\url{http://odds.cs.stonybrook.edu/satimage-2-dataset/}
\bibitem{c14}Lichman, M. (2013). UCI Machine Learning Repository [http://archive.ics.uci.edu/ml]. Irvine, CA: University of California, School of Information and Computer Science.

\end{thebibliography}
\end{document}